\documentclass[sigconf]{acmart}

\AtBeginDocument{%
  }

\setcopyright{rightsretained}
\copyrightyear{2026}
\acmYear{2026}
\acmDOI{}

\acmConference[CARS@RecSys'26]{Workshop on Context-Aware Recommender Systems}%
{September 28, 2026}{Minneapolis, MN}
\acmISBN{}
\usepackage{dsfont}
\usepackage{amsmath}
\usepackage{booktabs}
\usepackage{multirow}

\begin{document}

\title[SAGA: Structure-Attended Generative Action Embedding Model]{SAGA: Structure-Attended Generative Action Embedding Model that encodes Multi-Surface User Action Sequences}

\author{Tsz Fung Pang}
\orcid{0009-0009-7529-3171} 
\affiliation{%
  \institution{PayPal AI}
  \city{San Jose}
  \state{CA}
  \country{USA}
}
\email{tspang@paypal.com}

\author{Po Jen Chen}
\affiliation{%
  \institution{PayPal AI}
  \city{San Jose}
  \state{CA}
  \country{USA}
}
\email{pojchen@paypal.com}

\author{Nimish Ronghe}
\affiliation{%
  \institution{PayPal AI}
  \city{San Jose}
  \state{CA}
  \country{USA}
}
\email{nronghe@paypal.com}

\author{Farhad Farahani}
\affiliation{%
 \institution{PayPal AI}
 \city{San Jose}
  \state{CA}
  \country{USA}
}
\email{ffarahani@paypal.com}

\author{Bo Zhang}
\affiliation{%
 \institution{PayPal AI}
 \city{San Jose}
  \state{CA}
  \country{USA}
}
\email{bzhang7@paypal.com}

\renewcommand{\shortauthors}{Pang et al.}

\begin{abstract}
  Prior embedding models for sequential recommendation typically operate within a homogeneous action space, limiting their ability to capture cross-surface behavioral signals spanning distinct behavioral domains. We present SAGA, a generative action embedding model that encodes multi-surface user interaction sequences across a Financial Service organization's ecosystems, from checkout, peer-to-peer (P2P) transactions, in-app engagement, email to account actions, into a unified user representation for downstream recommendation tasks. Central to SAGA is a per-field tokenization schema that decomposes each action event into multiple field-level tokens (e.g. product, interaction, surface), enabling field-level attention and per-field training objectives that fused single-token approaches cannot support. Through an offline ablation study on loss formulation, tokenization granularity and training data scope, we isolate the contribution of each design choice. A downstream model integrated with SAGA-generated user embeddings delivers the strongest overall click and conversion lift across diverse downstream touchpoints, compared to all ablated and alternative architectures. 
\end{abstract}

\ccsdesc[500]{Information systems~Recommender systems}

\keywords{recommendation system, sequential recommendation, personalization, financial services, embedding model, action tokenization}

\begin{teaserfigure}
  \centering
  \includegraphics[width=\textwidth]{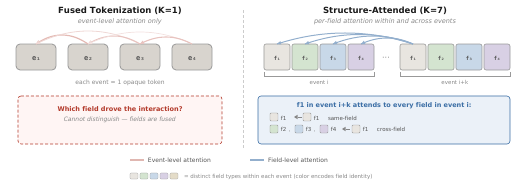}
  \caption{Structure-attended action sequences (right, $K{=}7$) enabled per-field attention. Fusing all fields into a single token per event (left, $K{=}1$) limits attention to event-level granularity, losing fine-grained dependencies.}
  \label{fig:teaser}
\end{teaserfigure}


\maketitle

\begin{figure*}[t]
  \centering
  \includegraphics[width=\textwidth]{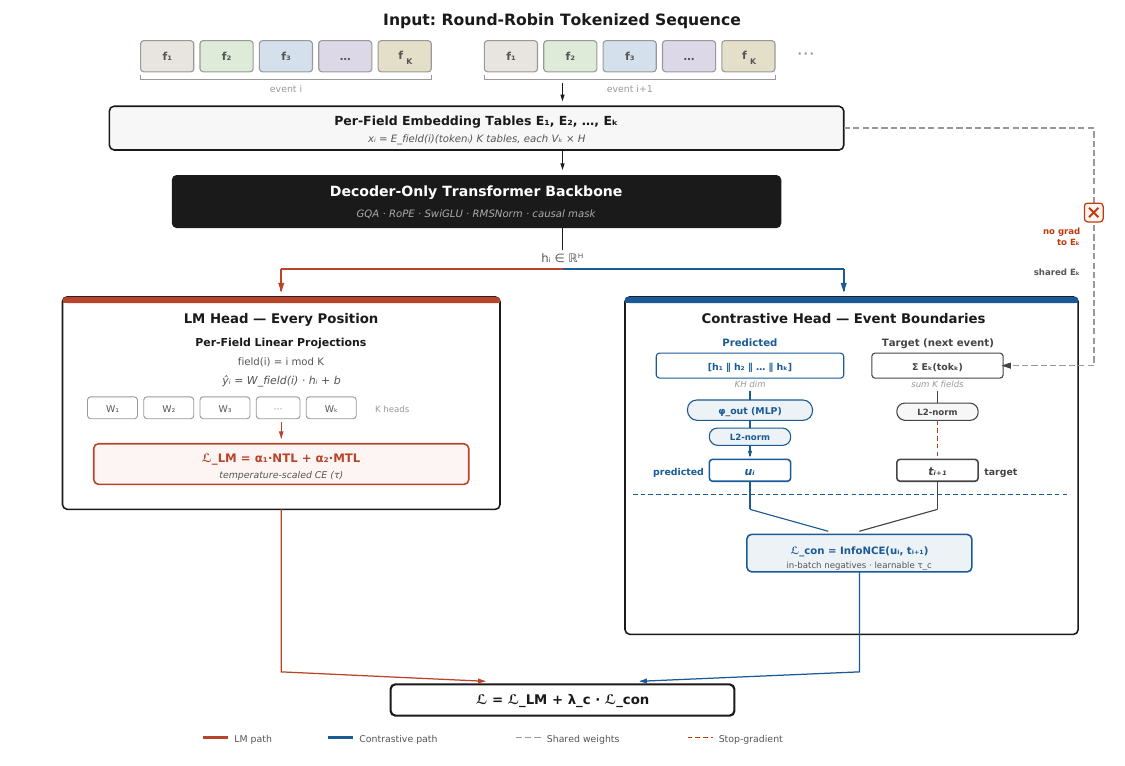}
  \caption{\textit{SAGA} dual-head architecture. A decoder-only transformer 
    with round-robin per-field tokenization ($K$ fields per event), 
    $K$ independent LM heads, and an event-boundary contrastive head 
    with stop-gradient compositional targets.}
  \Description{Architecture diagram showing decoder-only transformer 
    with round-robin per-field tokenization, ($K$ fields per event), 
    $K$ independent LM heads, and an event-boundary contrastive head 
    with stop-gradient compositional targets.}
  \label{fig:architecture}
\end{figure*}

\section{Introduction}
With the advancement of large language models, generative recommendation has attracted increasing attention as an alternative to conventional deep learning based recommender systems \cite{rajput2023recommender, zhai2024actions, chen2025pinfmfoundationmodeluser, xue2026generative}.  Traditional deep learning based
recommender pipelines often consist of multiple stages, such as retrieval and ranking, where discriminative neural network models are trained to predict outcomes such as clicks, conversions, or watch time. Generative recommendation reframes the problem by modeling user behavior as a sequence generation task, predicting the next action or item rather than directly estimating a task-specific response probability. This paradigm is expected to simplify cascaded recommendation pipelines, better leverage sequential behavioral data, and benefit from the scaling properties observed in large language models \cite{hou2026survey}.

Large-scale organizations tend to build specialized recommender systems to tackle recommendation problems independently and in silos as business problems expand and cannot be contained using a single model. A Financial Service (FS) company could have a need to personalize recommendations across different surfaces, notably marketing message outlets on App, Email and Push Notifications. E-commerce and FinTech companies also have a need to optimize checkout experience based on the upcoming user intent. Nonetheless, the independently built recommenders lack the synergy that would come from jointly learning across surfaces, and they cannot maintain consistency in the recommendation output. At best, data from other surfaces can be engineered and aggregated as features for a certain independent recommender model, but this model development practice is one-off and one-way. Upon development of recommender system for a new surface, there is often substantial duplication of feature engineering effort and the resulting machine learning model built is only applicable to its specialized problem and surface, leaving no shareable backbone that can be reused for different downstream tasks, as in the case of natural language and computer vision foundation models. 

Foundation models for sequential recommendation have gained traction in the industry \cite{chen2025pinfmfoundationmodeluser, zhai2024actions, ostroukhov2026pragmarevolutfoundationmodel, katariya2025fintrectransformerbasedunified}. These models pre-train on large-scale action sequences to learn reusable user representations, yet they typically operate within a single platform or a relatively homogeneous action space, leaving cross-surface learning across heterogeneous behavioral domains largely unexplored.

To alleviate the pain points of conventional model-building and fill in the gaps in sequential recommendation, we propose the \textbf{S}tructure-\textbf{A}ttended \textbf{G}enerative \textbf{A}ction  (\textbf{``SAGA''}) Model, an embedding model that encodes a user's historical action sequence on a structured attribute level across multiple surfaces, including Branded Checkout transactions, P2P transactions, engagement activities with the App, Push Notifications and Email, into a universal user representation that can be useful for multiple downstream tasks and surfaces. Structure-attended action sequence modeling enables per-field attentions across action events, which recovers the loss of fine-grained attention in fused tokenization where each action event is fused as one token, as illustrated in Figure \ref{fig:teaser}.

To generate a holistic user representation capable of powering a variety of downstream tasks, the embedding model should be pre-trained with multi-surface data of different behavioral domains and distributions, which creates challenges of negative transfer and joint optimization. We are addressing these challenges by aggregating multi-surface action data of heterogeneous distributions into a sequence of unified structured schema. Our key contributions are:
\begin{enumerate}
    \item Round-robin tokenization strategy that enables field-level attention across heterogeneous action attributes. 
    \item Dual-head training objective that exploits per-field structure by combining autoregressive field prediction with event-boundary contrastive learning.
    \item Results validating that SAGA unlocks the value of cross-surface pre-training, where diverse data sources become complementary rather than conflicting.
\end{enumerate}
Our proposed architecture is presented in Figure \ref{fig:architecture}. Our results show that SAGA is able to generate user representations transferable to a range of downstream tasks, outperforming other variants consistently. 


\begin{figure*}[t]
  \centering
  \includegraphics[width=\linewidth]{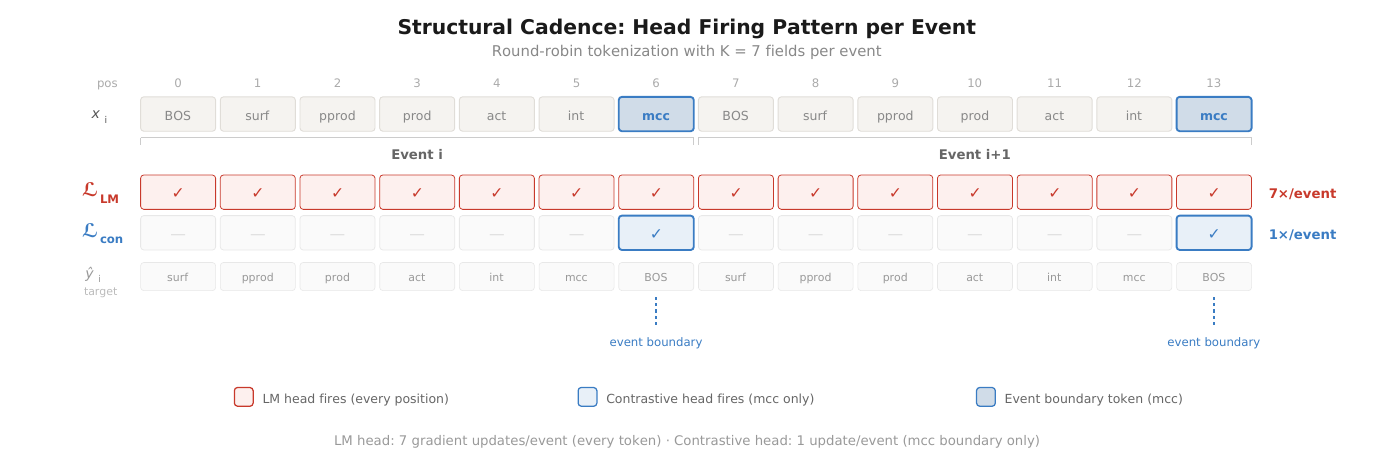}
  \caption{%
    Structural cadence of the dual-head firing pattern across two consecutive events. 
    The LM head $\mathcal{L}_{\mathrm{LM}}$ computes a next-token prediction loss at every position
    ($7{\times}$ per event), while the contrastive head $\mathcal{L}_{\mathrm{con}}$
    fires only at the \texttt{mcc} event-boundary token ($1{\times}$ per event). 
    The bottom row ($\hat{y}_i$) shows the shifted prediction targets for the LM head.
  }
  \label{fig:structural-cadence}
\end{figure*}

\section{Related Work}

Our study contributes to three streams of literature: generative recommendation, financial foundation models and action event tokenization. We study how an embedding model can learn reusable representations from heterogeneous financial and digital customer activities for downstream recommendation tasks.

\subsection{Generative Recommendation}

First, our work contributes to the growing literature on generative recommendation. Existing studies on generative recommendation can be broadly grouped into three approaches. The first directly models user behavior sequences for next-token or next-action prediction. Representative examples include HSTU \cite{zhai2024actions} and OneRec \cite{deng2025onerec}, which show the effectiveness of sequence-based architectures for large-scale recommendation. The second approach jointly incorporates sequential and non-sequential features into a unified model \cite{zeng2025interformer, hou2026kunlun, zhang2026onetransunifiedfeatureinteraction}. This design combines the strength of sequence modeling with rich contextual, user-level, and item-level features. The third approach separates the foundation model from downstream task-specific models. PinFM \cite{chen2025pinfmfoundationmodeluser} and Spotify's generalized foundation model \cite{fazelnia2025generalized}, for example, learn reusable representations from large-scale user behavior data and apply them to multiple recommendation tasks. This approach improves flexibility and can be more cost-efficient than training separate large models for each downstream objective. Our work is most closely related to the third direction, since SAGA learns reusable customer representations that can support multiple recommendation tasks.

A key difference is that SAGA targets a more heterogeneous multi-surface setting. Prior models often focus on a relatively homogeneous action space or a single platform \cite{zhai2024actions, chen2025pinfmfoundationmodeluser}. Their designs are effective for consistent action spaces, but are less directly suited to settings where checkout events, P2P payments, app sessions, web visits, emails, and push notifications carry different meanings and feature structures. Relatedly, ActionPiece represents user actions as collections of contextual feature tokens \cite{hou2025actionpiececontextuallytokenizingaction}. Building on this idea, SAGA adopts a unified event schema that preserves surface-specific information while modeling all actions in a shared sequence.

CL4SRec proposed a combination of contrastive learning with next item prediction in sequential recommendation \cite{xie2021contrastivelearningsequentialrecommendation}. CL4SRec's contrastive learning is achieved by comparing representations encoded from the augmented sequence (via random crop, mask, or reorder) of the same history. Their contrastive learning is motivated by post-augmentation view consistency of the sequence representation. SAGA differs from this construction and motivation. SAGA's contrastive head is tuned to generalize over next event prediction, while the model's next-token prediction operates on event fields. The contrastive objective is constructed as an in-batch comparison between positive and negative event embeddings.

\subsection{Financial Foundation Models}

Our work contributes to the emerging literature on financial foundation models. Recent studies show that large-scale financial histories can be used to learn transferable representations. PRAGMA models multi-source banking histories with key-value-time tokenization and masked modeling for tasks such as credit scoring, fraud detection, lifetime value prediction, engagement prediction, and recommendation \cite{ostroukhov2026pragmarevolutfoundationmodel}. TransactionGPT studies consumer payment trajectories using a 3D Transformer for transaction generation, anomaly detection, and representation learning \cite{dou2025transactiongpt}. nuFormer learns transaction embeddings from textual, numerical, and categorical attributes to improve recommendation at Nubank \cite{braithwaite2025your}. The Open Banking Foundational Model further shows that combining structured transaction attributes with textual descriptions improves generalization across institutions and data-scarce settings \cite{polleti2025openbankingfoundationalmodel}.

Compared with these studies, SAGA has a broader scope and a different objective. Prior financial foundation models mainly focus on banking histories, payment trajectories, or transaction-level representation learning. SAGA instead jointly models financial and non-financial customer activities, including checkout, P2P, App, Web, Email, and Push Notification interactions. This allows us to evaluate how diverse digital and communication signals complement financial behavior in downstream recommendation tasks.

\subsection{Action Event Tokenization}
Prior work has studied a variety of tokenization approaches for action event data. HSTU tokenizes each action event into two tokens, namely product and interaction \cite{zhai2024actions}. PinFM fuses the action event fields into a single embedding such that the generative backbone operates with one token for each event \cite{chen2025pinfmfoundationmodeluser}. ActionPiece decomposes item attributes into tokens using byte-pair encoding, a.k.a. context-dependent split-and-merge tokenization \cite{hou2025actionpiececontextuallytokenizingaction}. However, existing autoregressive approaches sacrifice field-level granularity through fusion or hard feature selection, while the per-attribute approach that preserves field identity (PRAGMA) operates in an encoder setting that precludes causal generation. No prior work combines per-field tokenization with autoregressive modeling in a deterministic token ordering that enables both field-level attention and constrained, surface-conditioned generation.

SAGA achieves this by decomposing each action event into seven fields, including six event attributes and a separator token ($K{=}7$), in contrast to PinFM($K{=}1$) and HSTU ($K{=}2$). SAGA also adopts a per-field tokenization approach in autoregressive modeling by enforcing a round-robin order across the fields, extending the tokenization design from the encoder architecture in PRAGMA. Autoregressive modeling provides the theoretical benefit of (a) next event field-specific vocabulary prediction, (b) structured event factorization, rather than implicit learning of event field joint distribution, and (c) constrained generation conditioned on specific surfaces. ActionPiece provides an orthogonal approach in tokenizing items with unordered variable-length attributes, whereas SAGA tokenizes each event into a fixed number of fields in a deterministic round-robin order. Only (a) is realized in our per-field LM heads (Section \ref{sec:dual-head}), while (b) and (c) are theoretical properties that we leave to future evaluation. 

To the best of our knowledge, SAGA is the first work to apply per-field round-robin tokenization in autoregressive modeling in the domain of recommender systems. This per-field autoregressive design enables SAGA to serve as a reusable embedding model across heterogeneous surfaces, as the field-level structure preserves surface-specific information that fused representations discard.



\begin{table}[H]
\centering
\caption{Notation for training objectives.}
\label{tab:notation}
\begin{tabular}{@{}cp{0.72\linewidth}@{}}
\toprule
\textbf{Symbol} & \textbf{Definition} \\
\midrule
$\mathbf{x}_{1:T}$  & Input token sequence of length $T$ \\
$m$                  & Number of tokens in the training sequence \\
$K$                  & Number of fields per event (${=}\,7$) \\
$s$                  & Stride for multi-token prediction ($s{=}K$ in round-robin mode) \\
$\mathbf{h}_i$       & Hidden state at position $i$ from the transformer backbone \\
$\hat{\mathbf{y}}_i$ & Logit vector at position $i$, produced by the LM head \\
$z_i$                & Ground-truth label at position $i$ \\
$\ell(\cdot,\cdot)$  & Per-position loss (SoftCE or CrossEntropy), operating on logits \\
$\tau$               & Temperature for SoftCE (default: 0.07) \\
$\tau_c$             & Learnable temperature for contrastive head (init: 0.07) \\
$\lambda_c$          & Contrastive loss weight (default: 0.5) \\
$\alpha_{\text{ntl}},\;\alpha_{\text{mtl}}$
                     & LM loss component weights (defaults: 1.0, 1.0) \\
\bottomrule
\end{tabular}
\end{table}
\section{Methodology}
\label{sec:method}
We pre-trained our embedding model and extracted the last layer hidden state as the frozen user embedding to be used as the downstream model input. We focus on evaluating the impact of design choices on the representation quality, so fine-tuning is out of the scope of this paper. The notations used in the paper are defined in Table \ref{tab:notation}.

\subsection{Event Attributes as Tokens} 
\paragraph{\textbf{Data Construction.}} We constructed a proprietary financial service dataset consisting of user actions aggregated from branded checkout transactions, peer-to-peer (P2P) transactions, engagement with App, push notifications, and emails, and account management. Each user action from different surfaces is featurized into a unified structured schema such that the user action sequences are formed by a set of action events of identical schema. Events are ordered chronologically to form a user action sequence.

\paragraph{\textbf{Unified Event Schema.}} We transform each user action event into six standardized attributes, including (1) surface, (2) parent product, (3) product, (4) action intent, (5) interaction and (6) merchant code categorization. The transformations are done by a predetermined lookup approach. \textit{Surface} represents the source of the user action event. It suggests the device of the checkout transaction or the app touchpoints that user interacted with. \textit{Parent product} and \textit{product} are a two-level hierarchy category of the financial products offered by a FS company, led by the examples of branded checkout and credit cards. \textit{Action intent} represents the granular user intent with respect to a particular action, such as cross border purchase or credit card sign up. \textit{Interaction} refers to the depth of user interactions, which identify impression, click and conversion into separate buckets. Transaction-typed data are assigned into different buckets of transaction amounts. \textit{Merchant category code (mcc)}  is an exclusive attribute for transaction-typed data, with the hypothesis that transaction activities that require financial commitments contain the most signal for the user representation. 

\paragraph{\textbf{Vocabulary construction.}} We create one distinct vocabulary list for each feature, each backed by a dedicated tokenizer. To encode the user representation with a small-sized transformer model, we intentionally remove item ID from the set of attributes and tokenizer, and only keep its mapped attributes. The vocabulary size  ranges from 9 to 239 tokens per field, totaling 378 tokens. Despite the compact vocabulary, round-robin composition yields a theoretical event space exceeding $10^9$. Expressiveness thus emerges from combinatorial composition rather than vocabulary scale.

\subsection{Round-robin Tokenization}

\paragraph{\textbf{Mechanism.}} We expanded each event into seven tokens in a round-robin order across the sequence, including the six attributes described above and a separator token (denoted as BOS) prepended at the beginning of each event. Each event is tokenized into $K{=}7$ round-robin fields
    (\texttt{BOS}, \texttt{surf}, \texttt{pprod}, \texttt{prod}, \texttt{act}, \texttt{int}, \texttt{mcc}), as shown in Figure \ref{fig:structural-cadence}. 
    Our seven-field structure is an extension from HSTU's two-field interleaving sequence formation, standing in contrast to the single fused embedding per event in PinFM. The decoding also follows the round-robin order to predict the next field of each event one by one, by looking up their respective distinct vocabulary tables. In the implementation, the out-of-vocabulary logits in the softmax layer are masked with negative infinity such that the model can only predict tokens drawn from a given field, according to the current round-robin order. We adopted standard Rotary Position Embedding (RoPE) \cite{su2023roformerenhancedtransformerrotary} to inject positional encoding into the sequence. 

\paragraph{\textbf{Motivations.}} The major benefit of our design is to enable per-field attention and constrained generation. For the prediction of the next event, the transformer architecture can attend to the embeddings of all attributes of prior events, whereas event-based fused embedding loses information in aggregating each event into a single embedding without leveraging the granular attention across the attributes of each event. Moreover, since our embedding model is designed to serve multiple surfaces, with the seven-field design, we can constrain and condition the generation of the next user action event on a particular surface, e.g. branded checkout, such that transformer can predict the most likely user events that can be served in the selected surface. This allows the surface-conditioned generation of the user embedding and also generative recommendation.

\subsection{Dual Heads}
\label{sec:dual-head}

\paragraph{Autoregressive Predictions}
We adopt a dual head architecture for SAGA. One head is the standard autoregressive next-token prediction head used in language model (``LM head'') with causal masking. We adopt the next-token loss (NTL), which computes the softmax loss between the target token and predicted logits, and the multi-token loss (MTL), which extends the softmax loss to a window of ground truth tokens within a selected horizon, proposed by PinFM\cite{chen2025pinfmfoundationmodeluser}. The order of user actions has higher variance than languages, so the MTL is added to accommodate this nature of generative action model. Following the round-robin order, each of the six fields (excluding separator) contributes to the softmax loss by comparing the logits of the valid token exclusively for the field and the next ground truth token. Figure \ref{fig:structural-cadence} demonstrates the shifted prediction target for LM head.

For each position, we compute the softmax loss as follows:
\begin{equation}
    \ell(\hat{\mathbf{y}}_i, z_j) = -\log \frac{\exp\bigl(\hat{y}_{i,z_j} / \tau\bigr)}{\sum_{v \in \mathcal{V}_f} \exp\bigl(\hat{y}_{i,v} / \tau\bigr)}
\end{equation}
where $\hat{y}_{i}$ is the logit at position $i$ and $\tau=0.07$. The NTL is formulated as below: 
\begin{equation}
    \mathcal{L}_{\text{NTL}} = \frac{1}{m-1} \sum_{i=1}^{m-1} \ell(\hat{\mathbf{y}}_i, z_{i+1})
\end{equation}
where LM-head-produced logit $\hat{y}_{i}$ are compared directly with the ground truth next token $z_{i+1}$. MTL loss is adjusted for the round-robin tokenization such that each predicted logit is compared against the ground truth token in the \textit{same} field for the next $k$ events: 
\begin{equation}
    \mathcal{L}_{\text{MTL}} = \frac{1}{|\mathcal{S}|} \sum_{i=1}^{m-1} \sum_{k=1}^{L'} \mathds{1}[i + k \cdot s \leq m] \; \ell(\hat{\mathbf{y}}_i, z_{i + k \cdot s}),
\end{equation} 
where $s{=}K{=}7$ is the stride in round-robin mode, predicting a field occurring within the $k$ events ahead, $L'$ is the multi-token prediction horizon, and $|\mathcal{S}|$ is the number of valid $(i, k)$ pairs satisfying $i + k \cdot s \leq m$.

\paragraph{Event-based Contrastive Head}
We theorize that training with the contrastive loss on the event-based embeddings will improve the action-centric user representation. We construct the event-based embedding prediction at the end of the event boundary by aggregating all $K$ hidden states of the prior event into a single event embedding via an MLP projection layer: 
\begin{equation}
    \mathbf{u}_i = \phi_{\text{out}}\!\left(\bigl[\mathbf{h}_{iK}^{(1)} \| \mathbf{h}_{iK+1}^{(2)} \| \cdots \| \mathbf{h}_{iK+K-1}^{(K)}\bigr]\right), 
\end{equation}
where $\phi_{\text{out}}:\mathbb{R}^{KH} \rightarrow \mathbb{R}^{H}$ is a post-norm two-layer MLP projection and $i$ indexes events here. 

Then we compute the InfoNCE loss between the projected event embedding against a target event embedding, which is constructed by element-wise addition of the embedding of 7 fields of the ground truth event in the next position: 
\begin{equation}
    \mathbf{t}_{i+1} = \mathrm{normalize}\!\left(\texttt{sg}\!\left[\sum_{f=1}^{K} \mathbf{e}_f\!\left(x_{(i+1)K + f - 1}\right)\right]\right)
\end{equation}
where $\mathbf{e}_f(\cdot)$ is the embedding lookup for field $f$ of the next event, and $\text{sg}$ refers to stop-gradient. Stop-gradient is added to the construction of target embedding such that the loss only backpropagates via the predicted event embedding and transformer architecture back to the embedding lookup layer, preventing representation collapse. 

The InfoNCE is computed between the in-batch negative samples and the constructed ground truth target embedding: 
\begin{equation}
    \mathcal{L}_{\text{con}} = -\frac{1}{|\mathcal{P}|} \sum_{i \in \mathcal{P}} \log \frac{\exp\bigl(\mathbf{u}_i^\top \mathbf{t}_{i+1} / \tau_c\bigr)}{\sum_{j=1}^{N} \exp\bigl(\mathbf{u}_i^\top \mathbf{t}_j / \tau_c\bigr)},
\end{equation}
where $\mathcal{P}$ is the set of positive interaction events, $N$ is the size of the in-batch negative samples. 

The loss is fired at the event boundary, i.e. every 7 tokens in the sequence, as demonstrated in Figure \ref{fig:structural-cadence}.

\paragraph{Combined Objectives}
The final training objective is a weighted sum of the NTL, MTL and contrastive loss: 
\begin{equation}
    \boxed{
\mathcal{L}_{\text{total}} = \underbrace{\alpha_{\text{ntl}} \cdot \mathcal{L}_{\text{NTL}} + \alpha_{\text{mtl}} \cdot \mathcal{L}_{\text{MTL}} }_{\mathcal{L}_{\text{LM}}} \;+\; \lambda_c \cdot \mathcal{L}_{\text{con}}
},
\end{equation}
where $\alpha_{\text{ntl}}=\alpha_{\text{mtl}}=1$ and $\lambda_c=0.5$.

\subsection{Model Design Choices}

\paragraph{Model Architecture.}
We adopted the decoder-only transformer with 12 layers, hidden dimension $H{=}512$, 8 attention heads with 4 key-value heads, as the backbone. This architecture is chosen for two reasons. First, as similar performance is observed between GPT-2 and HSTU architecture for recommendation tasks \cite{chen2025pinfmfoundationmodeluser}, we focus our study on tokenization and training objectives and fix the transformer architecture. Second, we can adopt compute-efficient modules, including Grouped-Query Attention (GQA), SwiGLU, RMSNorm, which are unavailable in the HSTU architecture. The context window is 1024 tokens, corresponding to approximately 146 events under round-robin tokenization ($K{=}7$ tokens per event). The model has approximately 37 million parameters. 

\paragraph{Training Data} All experimental variants are trained with the same set of user data spanning across 2 years. User action event sequences are split into non-overlapping chunks. For variants with different numbers of events in a sequence, sequences are split into chunks of different length, but all model variants are still trained with all chunks of the selected users. 

\paragraph{Embedding Extraction.}
We extracted the last layer hidden state following the input of the entire user sequence as the user embedding. For the scope of this paper, we froze the weights of SAGA, and also utilized the user embedding as the input of downstream models, which were trained and evaluated using the user embedding generated from a frozen SAGA.

\begin{table}[t]
\centering
\caption{Ablation study variants.  \textit{SAGA} (the proposed model) uses $K{=}7$ fields per event, dual-head pre-training ($\mathcal{L}_{\text{NTL}} + \mathcal{L}_{\text{MTL}} + \lambda_c \mathcal{L}_{\text{con}}$), and all interaction surfaces.  Each row modifies exactly one dimension; all other settings are held constant.  \textnormal{\textit{Set~A}} fixes events; \textnormal{\textit{Set~B}} fixes the token budget.}
\label{tab:ablation}
\small
\begin{tabular}{@{}cllcl@{}}
\toprule
\textbf{Abl.} & \textbf{ID} & \textbf{Variable changed} & \textbf{$K$} & \textbf{Training data} \\
\midrule
\multirow{3}{*}{\rotatebox[origin=c]{90}{\scriptsize Objective}}
  & A1 & LM heads only ($\lambda_c{=}0$)         & 7 & All \\
  & A2 & Contrastive only ($\alpha_{\text{ntl}}{=}\alpha_{\text{mtl}}{=}0$) & 7 & All \\
  & A3 & \textit{SAGA} (dual-head)                & 7 & All \\
\addlinespace
\multirow{3}{*}{\rotatebox[origin=c]{90}{\scriptsize Tok (A)}}
  & B1\textsubscript{A} & Fused-key, 146 evt            & 1 & All \\
  & B2\textsubscript{A} & Interleaved, 146 evt           & 2 & All \\
  & B3\textsubscript{A} & \textit{SAGA}, 146 evt           & 7 & All \\
\addlinespace
\multirow{3}{*}{\rotatebox[origin=c]{90}{\scriptsize Tok (B)}}
  & B1\textsubscript{B} & Fused-key, 1024 tok            & 1 & All \\
  & B2\textsubscript{B} & Interleaved, 1024 tok           & 2 & All \\
  & B3\textsubscript{B} & \textit{SAGA}, 1022 tok           & 7 & All \\
\addlinespace
\multirow{3}{*}{\rotatebox[origin=c]{90}{\scriptsize Data}}
  & C1 & ---                                       & 7 & Engagement \\
  & C2 & ---                                       & 7 & + Transactions \\
  & C3 & \textit{SAGA}                              & 7 & All surfaces \\
\bottomrule
\end{tabular}
\end{table}

\section{Offline Experiments}
\label{sec:offline}

We implemented an offline ablation study in a downstream task that requires ranking of in-App product recommendations. The study includes (1) comparison of LM-only and contrastive-only, and dual-head structure, (2) comparison of round-robin tokenization ($K{=}7$) against PinFM-style fused-key event embedding ($K{=}1$) and HSTU-style interleaved tokenization ($K{=}2$), (3) contributions of model trained with single-surface, few-surface, and all-surface data. All ablation variants are summarized in Table \ref{tab:ablation}. 

 We retrieved the frozen embeddings of all variants to train ranker models for downstream marketing message ranking tasks in the App. The downstream models are built with ESMM objective \cite{ma2018entirespacemultitaskmodel} and DCN-v2 architecture \cite{Wang_2021} and take production features as input parameters along with optional frozen embeddings as additional input. The no-embedding baseline only utilizes production features. All downstream models use identical architecture and hyperparameters across variants. We reported the results in three different touchpoints (``TP'') in the App. The metrics are offline NDCG@3, where positive labels indicate clicks and conversions. The values are relative \% improvement over no-embedding baseline model.  
Note that touchpoint-C is a low-engagement touchpoint where positive signals are sparse, and the signal-to-noise ratio is already low. Conversion lifts are not reported due to insufficient conversion volume. All rankers with SAGA variant embeddings produce negative click lift relative to the no-embedding baseline. We bold the least negative result as it indicates the least interference. We theorize that SAGA's behavioral embedding that encodes user's general interaction patterns may easily interfere with the already-weak signal in this touchpoint. The model variant that consistently radiates the least noise shall win. 

\subsection{Dual Head vs Single Head}


\begin{table}[t]
\caption{Training objective ablation. Values are relative \% improvement in NDCG@3 over no-embedding baseline.}
\label{tab:ablation-objective}
\resizebox{\columnwidth}{!}{%
\setlength{\tabcolsep}{3pt}
\begin{tabular}{ll*{6}{c}}
\toprule
& & \multicolumn{2}{c}{\textbf{TP-A}} & \multicolumn{2}{c}{\textbf{TP-B}} & \multicolumn{2}{c}{\textbf{TP-C}} \\
\cmidrule(lr){3-4} \cmidrule(lr){5-6} \cmidrule(lr){7-8}
\textbf{ID} & \textbf{Objective} & Clk & Conv & Clk & Conv & Clk & Conv \\
\midrule
A3 & \textbf{SAGA (dual-head)} & \textbf{+6.5} & +3.3 & \textbf{+1.7} & \textbf{+0.6} & \textbf{-1.4} & --- \\
A1 & LM only ($\lambda_c{=}0$) & +3.5 & +1.6 & +0.4 & +0.3 & -4.2 & --- \\
A2 & Contr.\ only ($\alpha_{\text{ntl}}{=}\alpha_{\text{mtl}}{=}0$) & +5.8 & \textbf{+3.9} & +0.5 & +0.4 & -8.6 & --- \\
\bottomrule
\end{tabular}%
}
\vspace{2pt}
\end{table}

We ablated the dual head training objectives to train SAGA with LM head only and contrastive head only. Table \ref{tab:ablation-objective} shows the dual-head SAGA has on aggregate outperformed single-head variants in terms of the lift in clicks and conversions. It also contributes to the least noise interference in touchpoint-C. Contrastive-only variant achieves greater lift in conversion on touchpoint-A but it also degrades the most on touchpoint-C. The results validated that the dual-head training objectives create user representation that can transfer most gain to a variety of downstream tasks.

The dual-head objective creates complementary learning signals: the LM heads learn fine-grained field-level transition dynamics, while the contrastive head learns distinguishable event-level representations. Neither objective alone captures both granularities.

\subsection{Action Event Tokenization}


\begin{table}[t]
\caption{Tokenization granularity ablation. \textit{Set~A}: fixed 146 events; \textit{Set~B}: fixed 1024-token budget.}
\label{tab:ablation-tokenization}
\resizebox{\columnwidth}{!}{%
\setlength{\tabcolsep}{3pt}
\begin{tabular}{llc*{6}{c}}
\toprule
& & & \multicolumn{2}{c}{\textbf{TP-A}} & \multicolumn{2}{c}{\textbf{TP-B}} & \multicolumn{2}{c}{\textbf{TP-C}} \\
\cmidrule(lr){4-5} \cmidrule(lr){6-7} \cmidrule(lr){8-9}
\textbf{ID} & \textbf{Tokenization} & $K$ & Clk & Conv & Clk & Conv & Clk & Conv \\
\midrule
\multicolumn{9}{l}{\textit{Set A --- Fixed 146 events:}} \\
B3\textsubscript{A} & SAGA (round-robin) & 7 & +6.5 & +3.3 & \textbf{+1.7} & \textbf{+0.6} & \textbf{-1.4} & --- \\
B1\textsubscript{A} & Fused (PinFM) & 1 & \textbf{+7.3} & \textbf{+4.5} & +0.6 & +0.4 & -3.9 & --- \\
B2\textsubscript{A} & Interleaved (HSTU) & 2 & +3.7 & +2.9 & -1.8 & -0.7 & -12.6 & --- \\
\midrule
\multicolumn{9}{l}{\textit{Set B --- Fixed 1024-token budget:}} \\
B3\textsubscript{B} & \textbf{SAGA, 146 evt} & 7 & \textbf{+6.5} & \textbf{+3.3} & \textbf{+1.7} & \textbf{+0.6} & \textbf{-1.4} & --- \\
B1\textsubscript{B} & Fused, 1024 evt & 1 & +6.4 & +3.0 & +0.9 & +0.5 & -16.0 & --- \\
B2\textsubscript{B} & Interleaved, 512 evt & 2 & +5.6 & +3.1 & -1.3 & -0.5 & -10.5 & --- \\
\bottomrule
\end{tabular}%
}
\vspace{2pt}
\end{table}
We carried out two sets of ablation study for tokenization granularity. In Set A, we fixed all variants to use 146 events, matching SAGA. In Set B, we fill up the 1024 token budget for other tokenization variants, where fused-key and interleaved design can accommodate up to 1024 and 512 events respectively.

From Table \ref{tab:ablation-tokenization}, SAGA has demonstrated consistently the highest performance gains in clicks and conversions across most touchpoints in both Set A and B compared to the other two tokenization designs. Fused-key achieved highest lifts on touchpoint-A in Set A  (+7.3\% vs +6.5\% and +4.5\% vs +3.3\%). In Set A, fused-key benefits from a more compact sequence (146 tokens vs. 1022 for SAGA), concentrating the model's attention budget on event-level temporal patterns. Its compressed event-level representations capture the user dynamics on the high-signal touchpoint. Nonetheless, the advantage does not transfer to the other two touchpoints, where its lifts drop to +0.6\% and +0.4\% on touchpoint-B and -3.9\% on touchpoint-C, exhibiting larger cross-touchpoint variance than SAGA. SAGA prevails as an embedding model which shall generate universally informative representations that are reusable across touchpoints. We attribute this result to the information bottleneck of fused tokenization in which compression of event attributes into a single embedding prioritizes the dominant signals at the expense of fine-grained field-level details needed for diverse downstream tasks.

When the token budget is equalized in Set B, SAGA at 146 events matches or exceeds fused-key with 1024 events and interleaved with 512 events, with $7\times$ fewer events within the same context window. Fused-key's advantage on touchpoint-A over SAGA evaporates (6.4\% vs 6.5\% and 3.0\% vs 3.3\%), while it degrades severely on touchpoint-C. This reinforces our hypothesis that per-field tokenization generates a more universally transferable representation. It also indicates that, at equal token budget, richer per-event tokenization is more effective than longer context with coarser representations. It can be explained by the fact that recent events carry stronger signal for predicting upcoming behavior. Encoding granular attributes of fewer recent events is more informative than aggregating compressed representations of more distant events.

Notably interleaved tokenization underperforms fused-key. We hypothesize that, despite fusion of granular event fields into one embedding, fused-key's learned sum-pooling implicitly reweights field contributions, approximating a soft attention over fields within each event. This partially recovers field-level signal. Interleaved tokenization, by contrast, hard-selects only product and interaction, permanently discarding surface, action intent, parent product, and MCC. The loss of attribute-level information cannot be recovered even with longer sequences.

\subsection{Multi-surface Data Enrichment}


\begin{table}[t]
\caption{Training data composition ablation. All variants use SAGA ($K{=}7$, dual-head).}
\label{tab:ablation-data}
\resizebox{\columnwidth}{!}{%
\setlength{\tabcolsep}{3pt}
\begin{tabular}{ll*{6}{c}}
\toprule
& & \multicolumn{2}{c}{\textbf{TP-A}} & \multicolumn{2}{c}{\textbf{TP-B}} & \multicolumn{2}{c}{\textbf{TP-C}} \\
\cmidrule(lr){3-4} \cmidrule(lr){5-6} \cmidrule(lr){7-8}
\textbf{ID} & \textbf{Training surfaces} & Clk & Conv & Clk & Conv & Clk & Conv \\
\midrule
C3 & \textbf{SAGA (All surfaces)} & \textbf{+6.5} & \textbf{+3.3} & \textbf{+1.7} & \textbf{+0.6} & \textbf{-1.4} & --- \\
C2 & Engage.\ + Transactions & +2.5 & +0.7 & -1.7 & -0.7 & -4.7 & --- \\
C1 & Engagement only & +3.1 & +1.4 & -0.9 & -0.3 & -5.6 & --- \\
\bottomrule
\end{tabular}%
}
\vspace{2pt}
\end{table}

Table \ref{tab:ablation-data} indicates that SAGA trained with all surfaces of data contributed to the greatest lifts in both click and conversion in all touchpoints. It is interesting to point out that the gain is not monotonic, as the addition of transaction data alone deteriorates the performance throughout most touchpoints. Transaction data is of low-frequency and strong signals, whereas engagement data is of high-frequency and weak signals. Adding only transactions creates a bimodal distribution that the model cannot reconcile. Adding all surfaces provides enough distributional diversity to learn cross-domain transfer. We interpret the results as the offsetting effects between cross-domain signal gain and negative transfer from other domains. Only SAGA trained with data of all surfaces achieved high enough cross-domain learning to offset the noise and interference in mixing the data, whereas the model variant trained with only engagement and transactions data suffers from the negative transfer from the mixture of data from two domains. 

\section{Online A/B Testing Results}
\begin{table}[t]
\centering
\caption{Online A/B test results. Values are relative \% lift over the control group (prior deep learning ranker).}
\label{tab:online-results}
\small
\begin{tabular}{@{}lcc@{}}
\toprule

\textbf{Touchpoint} & \textbf{Clk}& \textbf{Conv}\\
\midrule
TP-B & +5.05\%& +1.53\%\\
TP-C & $-$0.06\%& \\
\bottomrule
\end{tabular}
\end{table}

We also deployed SAGA in production A/B tests across touchpoints B and C. The treatment group received recommendations from a downstream model augmented with SAGA’s frozen embeddings, while the control group used the same ranking model without SAGA embeddings. The experiments evaluated the impact of SAGA embeddings on final clicks and conversions.

Table \ref{tab:online-results} summarizes the results. At touchpoint B, SAGA achieved a +5.05\% lift in click and a +1.53\% lift in conversion. At touchpoint C, SAGA resulted in a small 0.06\% decrease in click, which was not statistically significant. Overall, the A/B test results are consistent with the offline evaluation. They demonstrate that frozen embeddings learned through multi-surface pre-training with a structure-attended architecture can improve ranking performance on one surface without task-specific fine-tuning while maintaining performance on another surface.

\section{Future Work}
We acknowledge the finding from PinFM that fine-tuning will optimize the contributions of FMs towards downstream tasks \cite{chen2025pinfmfoundationmodeluser}, which is out of the scope of this study. As a continuation of this study, we propose fine-tuning the model variants against the downstream task by unfreezing the model weights, such that we can test the capacity of different model architectures. 

Round-robin tokenization also enables constrained generation that allows surface-specific generative recommendations, which we did not evaluate in the downstream tasks. It is a theoretical benefit from the architecture that should be explored at length in the follow-up study.  

We also fixed the model size and backbone, acknowledging the need to further validate the scaling law for structure-attended generative action models in the domain of recommender systems. 

\section{Conclusion}
We found that round-robin per-field tokenization ($K{=}7$) enabled structure-attended action sequences, which show the greatest performance improvements across downstream tasks in APP compared to ablated variants. Our ablation study shows that, given the same token budget, richer per-field tokenization generates more informative representation than event fused-key ($K{=}1$) and interleaved tokenization ($K{=}2$) with longer contexts and coarser event tokens. We also show that contrastive training objective and next token prediction objective are mutually complementary in the training of embedding model, and the combination of the two objectives is more effective than either alone. Lastly, our study demonstrates the benefit of cross-surface learning, compared to training user representation with single-domain data.

These results suggest that financial service organizations can build a single embedding model across heterogeneous customer touchpoints, replacing the current practice of surface-specific model development.




\bibliographystyle{ACM-Reference-Format}
\bibliography{references}

 
\appendix







\end{document}